# Analyzing Aviation Safety Narratives with LDA, NMF and PLSA: A Case Study Using Socrata Datasets

Aziida Nanyonga
School of Engineering and Information Technology
University of New South Wales
Canberra, Australia
a.nanyonga@adfa.edu.au

Graham Wild
School of Science
University of New South Wales
Canberra, Australia
g.wild@adfa.edu.au

***Abstract*—This study explores the application of topic modelling techniques Latent Dirichlet Allocation (LDA), Non-negative Matrix Factorization (NMF), and Probabilistic Latent Semantic Analysis (PLSA) on the Socrata dataset spanning from 1908 to 2009. Categorized by operator type (military, commercial, and private), the analysis identified key themes such as pilot error, mechanical failure, weather conditions, and training deficiencies. The study highlights the unique strengths of each method: LDA's ability to uncover overlapping themes, NMF's production of distinct and interpretable topics, and PLSA's nuanced probabilistic insights despite interpretative complexity. Statistical analysis revealed that PLSA achieved a coherence score of 0.32 and a perplexity value of -4.6, NMF scored 0.34 and 37.1, while LDA achieved the highest coherence of 0.36 but recorded the highest perplexity at 38.2. These findings demonstrate the value of topic modelling in extracting actionable insights from unstructured aviation safety narratives, aiding in the identification of risk factors and areas for improvement across sectors. Future directions include integrating additional contextual variables, leveraging neural topic models, and enhancing aviation safety protocols. This research provides a foundation for advanced text-mining applications in aviation safety management.**

## I. Introduction

Aviation narratives from publicly available datasets, such as Socrata, provide a wealth of textual information that captures incidents, operational details, and usage contexts [1]. These narratives often highlight critical details about the operators involved, whether they represent military entities like the U.S. Army or Navy, commercial airlines, or private operators. Understanding the categorization of operators in aviation reports is essential for incident analysis, resource allocation, and informed decision-making [2]. Military aviation, for instance, focuses on defence, investigation, training, and combat missions, involving specialised equipment and protocols that set it apart from civilian and private aviation [3]. In contrast, commercial airlines prioritise passenger and cargo services, adhering to strict safety standards and operational schedules while private aviation encompasses individual aircraft owners, charter services, and small-scale operators, often reflecting personal or recreational uses in their narratives[4].

Despite these distinctions, the textual summaries across these operator categories share overlapping linguistic features, making manual classification challenging and prone to errors. Automating the classification process through topic modelling offers a promising solution by uncovering latent themes within the narratives that correspond to operator categories. While significant research has explored large-scale aviation datasets for general incident analysis and safety reporting, limited attention has been paid to distinguishing specific operator categories such as military, commercial, and private aviation. The application of topic modelling techniques, including Latent Dirichlet Allocation (LDA), Non-Negative Matrix Factorization (NMF), and Probabilistic Latent Semantic Analysis (PLSA), to small-scale, narrative-focused datasets remains underexplored [5-7].

The primary objective of this study is to investigate the efficacy of these three topic modelling techniques in classifying aviation narratives into the categories of military, commercial airline, and private operators. This research focuses on analyzing narratives from the Socrata dataset, evaluating the coherence and perplexity of each technique, and identifying the linguistic patterns and themes. Accurate classification of these narratives holds practical significance across multiple domains. For military aviation, it enables better tracking of activities and compliance with defence protocols. For commercial airlines, it supports improved safety measures, maintenance practices, and operational efficiency. Finally, for private operators, it aids in incident analysis and policy development tailored to small-scale aviation.

In the subsequent sections of this paper, we will review related work, outline the methodology, present the experimental results, and provide a thorough discussion of our findings. Additionally, we will examine the implications of our research for aviation safety and propose potential directions for future studies.

## II. Related Work

Topic modelling has emerged as a critical tool for analyzing large-scale textual data, providing insights into hidden structures and themes within unstructured narratives. In aviation safety, it has been widely employed to identify trends, categorize incidents, and improve operational decision-making. Various studies have utilized topic modelling to analyze datasets such as NTSB narratives, highlighting their versatility and value in the field [6-8]. However, limited attention has been given to applying topic modelling to the Socrata dataset, which is the focus of this work.

Latent Dirichlet Allocation (LDA), introduced by Blei [9], has been extensively applied in diverse domains, including aviation, to uncover latent themes in textual data. For example, LDA has been employed to analyze FAA incident reports, extracting patterns related to human factors and mechanical failures [10]. Despite its utility, these analyses did not focus on differentiating operators, which constrains their applicability to targeted studies like this research. Similarly, Non-Negative Matrix Factorization (NMF), proposed by Lee and Seung [11], has emerged as a robust alternative for topic modelling, particularly when interpretability and sparsity are prioritized [7]. Although NMF has demonstrated potential in fields such as healthcare and e-commerce, its application in aviation remains underexplored. Studies have shown NMF and LDA to be effective in producing meaningful topics in small datasets, which aligns with this research's objectives [1, 12]

Probabilistic Latent Semantic Analysis (PLSA), introduced by Hofmann [9], has also been widely utilized in various domains to model co-occurrence data, including textual narratives. Its probabilistic framework has been effective for smaller datasets, making it suitable for this study's focus on the Socrata dataset. Nevertheless, PLSA's computational complexity and tendency to overfit have limited its broader adoption in aviation-related research [5]. Notably, Nanyonga et al. applied PLSA to analyze safety narratives from regional airlines, uncovering distinct operational patterns. However, their study did not extend to distinguishing between operator categories [5].

In the aviation domain, recent research has underscored the importance of operator-specific analysis. Military aviation narratives, for instance, emphasize operational risks unique to defense missions, such as mid-air refueling and formation flying. These risks differ significantly from the logistical and passenger safety priorities of commercial airlines. In contrast, private aviation narratives often highlight issues like recreational use, weather-related incidents, and less regulated maintenance practices. Despite the apparent distinctions, few studies have systematically explored these differences, leaving a significant gap in the literature.

This research aims to address this gap by building upon existing methodologies, employing LDA, NMF, and PLSA to classify narratives in the Socrata dataset into categories such as military, commercial airline, and private operators. Through a comparative analysis of these techniques, this study provides a nuanced understanding of their strengths and limitations in operator-specific applications. Moreover, this work contributes to the growing body of literature on aviation safety by emphasizing operator-specific insights, which are crucial for targeted interventions and policy development.

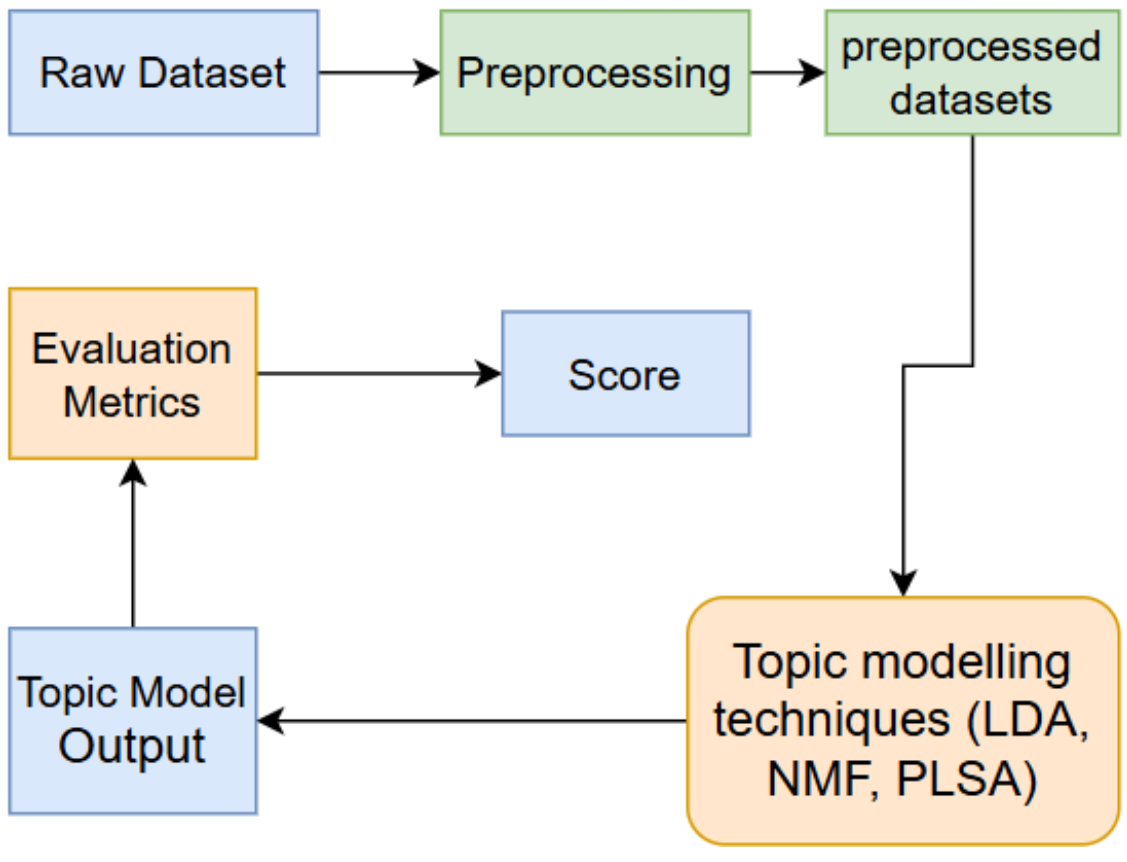

Fig. 1. Methodological framework

## III. Methodology

This section outlines the processes and techniques employed in this study to perform a comparative analysis of three prominent topic modelling techniques: pLSA, LDA, and NMF. The methodology includes data collection, preprocessing steps, the implementation of each modelling technique, and the evaluation metrics used to assess their performance.

### A. *Data Preprocessing*

The dataset used in this study is the Socrata Aviation Safety Dataset, which is publicly accessible and spans records from September 17, 1908, to June 8, 2009. The dataset contains 4,995 records, with fields that describe various circumstances of aviation incidents, with particular emphasis on the narrative fields outlining the circumstances of each event. This study focused on narratives that explicitly mention the operator type, categorized into military, commercial airlines, and private aviation.

### B. *Text Preprocessing*

The Socrata dataset required extensive preprocessing to ensure the quality and relevance of narratives for topic modelling. Initially, narratives were cleaned to remove special characters, numerical values, and extraneous symbols, creating a uniform text format. Tokenization was performed to break down the narratives into individual words, followed by the removal of stopwords using an aviation-specific stopword list to retain only meaningful terms. Lemmatization was then applied to standardize words to their base forms (e.g., “flying” to “fly”), reducing redundancy in the dataset. Narratives that explicitly mentioned operator types (military, commercial airline, or private aviation) were prioritized, while incomplete or ambiguous records were excluded. This preprocessing yielded three subsets of narratives corresponding to the operator categories, forming the basis for topic modelling. The implementation utilized tools such as NLTK for text processing and Pandas for dataset management.

### C. *Topic Modeling Procedure*

Following the preprocessing of the textual data, the next crucial step involves transforming the processed data into numerical features suitable for the application of topic modelling techniques. For this study, three well-established topic modelling methods LDA, NMF, and PLSA were employed. Each method was implemented separately, and topics were extracted from the dataset to reveal insights into the nature of aviation incidents involving operators. The entire process is illustrated in Fig. 1.

The preprocessed textual data were transformed into a Document-Term Frequency Matrix (DTF), which represents the frequency of each word across all narratives contained in the accident reports [10, 11]. This matrix serves as a structured numerical representation of the textual data, facilitating the application of the topic modelling techniques.

#### 1) *Latent Dirichlet Allocation (LDA)*

The first technique utilized in this study is LDA, a probabilistic generative model that assumes each document (aviation report) is a mixture of several topics, and each topic is a distribution over words. LDA has been widely used in natural language processing and is particularly effective for uncovering latent thematic structures in large text corpora. In the context of aviation safety, LDA has been used to analyze accident reports and identify recurring patterns in safety-related topics [6, 12]. This model works by inferring the topic distribution for each document and the word distribution for each topic, making it an ideal choice for analyzing the Socrata dataset, where the aim is to uncover hidden themes related to military, commercial airline, and private aviation operations. The implementation of LDA was carried out using the Gensim library, with the number of topics optimized based on coherence scores.

*2) Probabilistic Latent Semantic Analysis (pLSA)*

Probabilistic Latent Semantic Analysis (PLSA) is a probabilistic extension of Latent Semantic Analysis (LSA) that models the co-occurrence of words and documents using a probabilistic framework [9]. Unlike traditional methods, PLSA assigns each word in a document a latent topic, which is modelled through a probability distribution. In this study, PLSA was employed to uncover hidden topics in aviation safety reports by estimating the likelihood of topics within each document and words within each topic. This technique allows for more flexibility in modelling topic-word distributions and has shown promising results in other fields of text mining. The PLSA model was implemented using a custom Python package [13], and cross-validation techniques were applied to prevent overfitting and ensure the robustness of the results. Through PLSA, this research aimed to understand the underlying thematic patterns in aviation safety narratives, especially related to the differences between military, commercial, and private aviation operations.

*3) Non-negative Matrix Factorization (NMF).*

Another technique applied in this study is NMF, which decomposes the Document-Term Frequency Matrix into two lower-dimensional matrices: one representing topics and the other representing word distributions across topics [14]. NMF works by ensuring that the factorized matrices are non-negative, enhancing the interpretability of the results. This technique is particularly useful in cases with a smaller data set, as it highlights the most relevant terms and documents. In this study, NMF was implemented using Scikit-learn and was found to provide insightful topic extraction, especially when analyzing smaller subsets of narratives. NMF's strength lies in its ability to provide clearer, more interpretable topics that can help distinguish between operator categories, such as military, commercial, and private aviation. The number of topics was tuned iteratively to maximize the coherence and interpretability of the extracted topics.

### D. Evaluation Metrics

The effectiveness of the topic modelling techniques was evaluated using two key metrics: Perplexity and Coherence Score.

*1) Perplexity*

Perplexity measures the model's goodness of fit, specifically how well it predicts unseen documents based on the topics learned from the training data. Mathematically, perplexity is defined as the inverse probability of the test set normalized by the number of words. Lower perplexity values indicate better model performance, signifying that the model predicts the next word in the corpus with a higher probability [15, 16].

The formula for perplexity is as follows:

$$Perplexity(D) = \exp\left(-\frac{\sum_{d=1}^{M}\log p(D_d)}{\sum_{d=1}^{M} N_d}\right)$$

where:

$D_d$ is the $d-th$ document in the test set.

$N_d$ is the number of words in the $d-th$ document.

$p(D_d)$ is the likelihood of the document $D_d$ under the model.

*2) Coherence Score*

Coherence measures the semantic consistency of the topics by evaluating the degree of co-occurrence between words within a topic. Higher coherence scores indicate that the words associated with a topic are more likely to appear together in the corpus, reflecting the semantic relevance and interpretability of the topics [17].

The formula for coherence is as follows:

$$Coherence(T) = \sum_{i,j \in Top-N\ Words\ of\ Topic} Similarity(w_{i,}w_j)$$

where:

- $w_{i,} and\ w_j$ are words from the topic ($T$).
- Similarity measures how often the words $w_{i,} and\ w_j$ appear together in the corpus.

## IV. Results and Discussion

This section evaluates the outcomes of the topic modeling techniques LDA, PLSA AND NMF applied to Socrata dataset. The performance of the models was assessed using coherence scores and perplexity values, with tables summarizing the results and comparisons. Word clouds and topic distributions were generated to visually interpret the themes identified. Each model's strengths, weaknesses, and thematic relevance are discussed as follows.

TABLE I. Comparison of Topics Generated By PLSA, NMF, And LDA Models In Aviation Accident Analysis

| Topic Model | Key Themes | Common Words | Focus Areas |
|---|---|---|---|
| **PLSA** | Aircraft accidents, flight conditions, pilot actions | "plane", "crashed", "aircraft", "engine", "pilot", "landing", "weather" | Focus on accidents involving mechanical failures, weather, and pilot error. |
| **NMF** | Aircraft operations, emergencies, weather, and mechanical issues | "engine", "failure", "emergency", "fuel", "crashed", "runway" | Emphasizes mechanical failures, poor weather conditions, and flight emergencies. |
| **LDA** | Flight operations, weather conditions, accident causes | "flight", "weather", "pilot", "crashed", "terrain", "approach" | Highlights factors like weather, pilot decisions, and terrain-related causes of accidents. |

## A. Topic Distribution Across the Dataset

In this section, the topic distributions generated by three models; LDA, NMF, and PLSA are compared. Each model captures the underlying topics in the dataset differently due to their distinct mathematical approaches. The topic distribution across the dataset reveals distinct patterns when comparing these three models. LDA distributes words across multiple topics, with Topic 0 standing out as the most prominent, suggesting a balanced mixture of topics throughout the dataset. In contrast, NMF emphasizes Topic 3, indicating that this topic is the most representative of the dataset, with other topics contributing less to the overall content. PLSA, on the other hand, identifies Topic 6 as the dominant one, with most key terms concentrated on this topic. These differences in topic distribution reflect the unique modelling approaches of each algorithm, with LDA offering a more balanced topic spread, NMF focusing more on a single, defining topic, and PLSA pinpointing one specific topic as central to the dataset. The figures below further illustrate these distributions: Fig. 2 showcases the topic proportions for LDA, Fig. 3 for NMF, and Fig. 4 for PLSA.

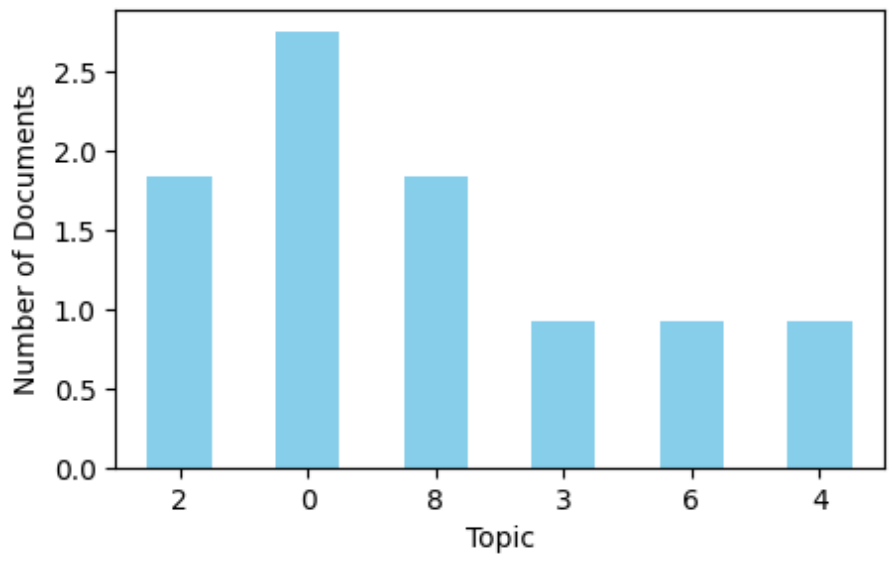

Fig. 2. Topic Distribution Using LDA model

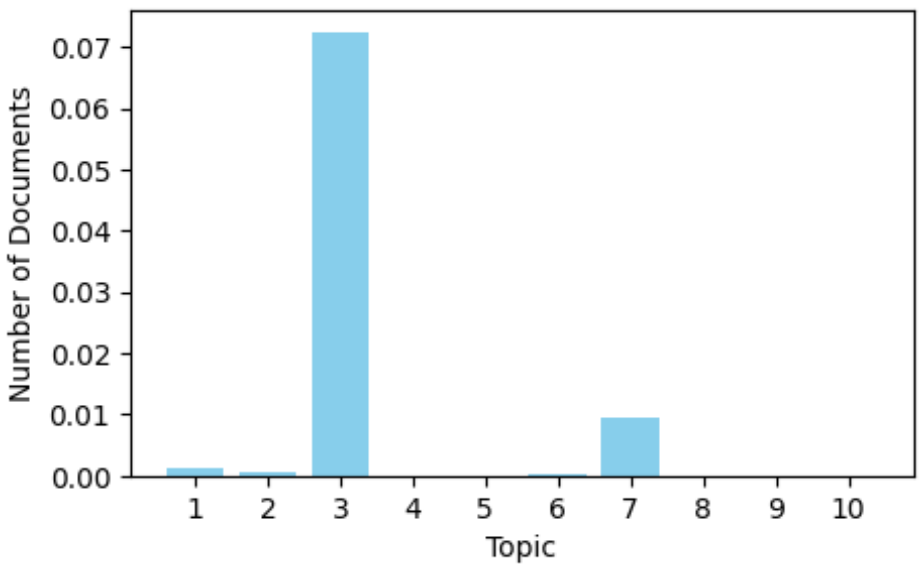

Fig. 3. Topic Distribution Using NMF model

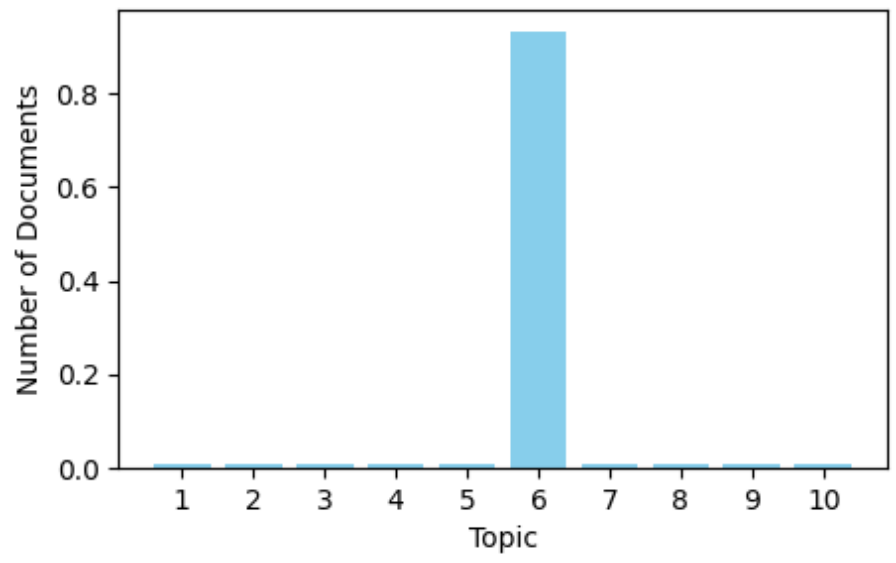

Fig. 4. Topic Distribution Using PLSA model

## B. Topic Extraction Using LDA

For each model, the key themes and common words are extracted to highlight the focus areas of the topics. To visualize the results, word clouds for each model are presented: Fig. 5 shows the word cloud for LDA, Fig. 6 for NMF, and Fig. 7 for PLSA. Table 1 summarizes the common words associated with each topic and the key themes of those topics.

The topics generated by PLSA focus heavily on specific accident scenarios, with prominent keywords such *as "plane"*, *"crashed"*, *"pilot"*, *"aircraft"*, and *"landing"*. For instance, Topics in PLSA highlights mechanical failures, pilot actions, and environmental factors that lead to accidents, as reflected in words like *"engine"*, *"control"*, *"weather"*, and *"pilot"*. The topics suggest a comprehensive view of aviation accidents, emphasizing mechanical breakdowns and pilot-related errors.

NMF's topics revolve around operational issues, weather conditions, and mechanical failures. Topic 1 emphasizes engine failure and emergencies, while Topic 2 focuses on terrain and mountainous areas, contributing to accidents. Words such as *"emergency"*, *"fuel"*, *"crashed"*, and *"runway"* point to mechanical failures, poor weather conditions, and flight emergencies, showcasing the technical and environmental aspects of aviation safety.

LDA's topics provide a broader perspective on flight operations, weather conditions, and the causes of accidents. Topic 1 in LDA focuses on flight conditions, with words like *"weather"*, *"pilot"*, and *"terrain"*, highlighting the influence of weather and pilot decisions in accidents. Other topics, such as Topic 3, emphasize mechanical failures, runway incidents, and the roles of pilots and crew in accidents. The LDA model captures a more comprehensive set of factors that contribute to aviation accidents.

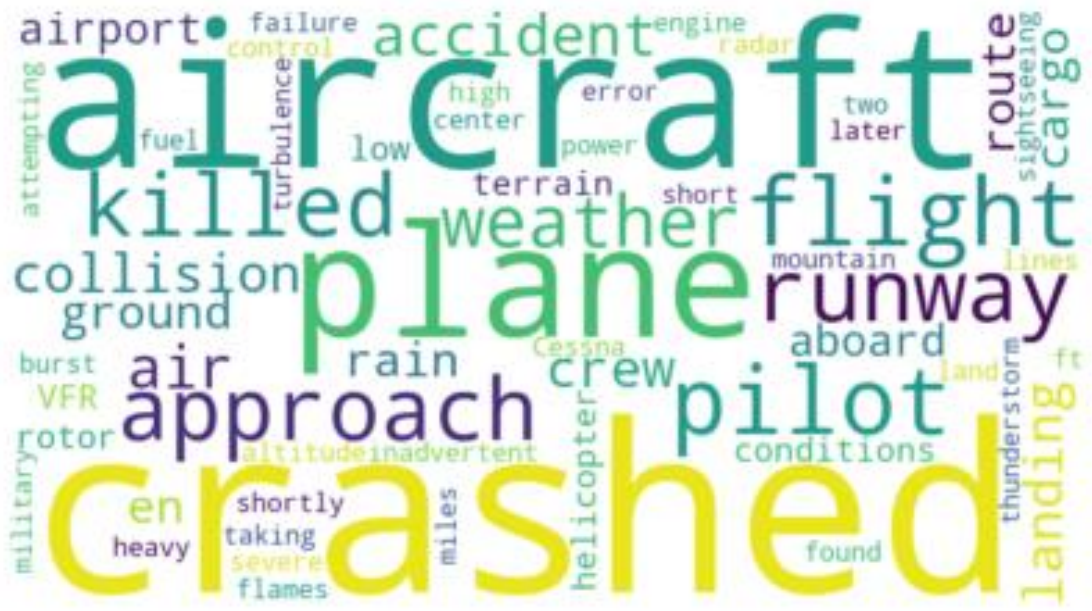

Fig. 5. Word Cloud of Topics for LDA Model

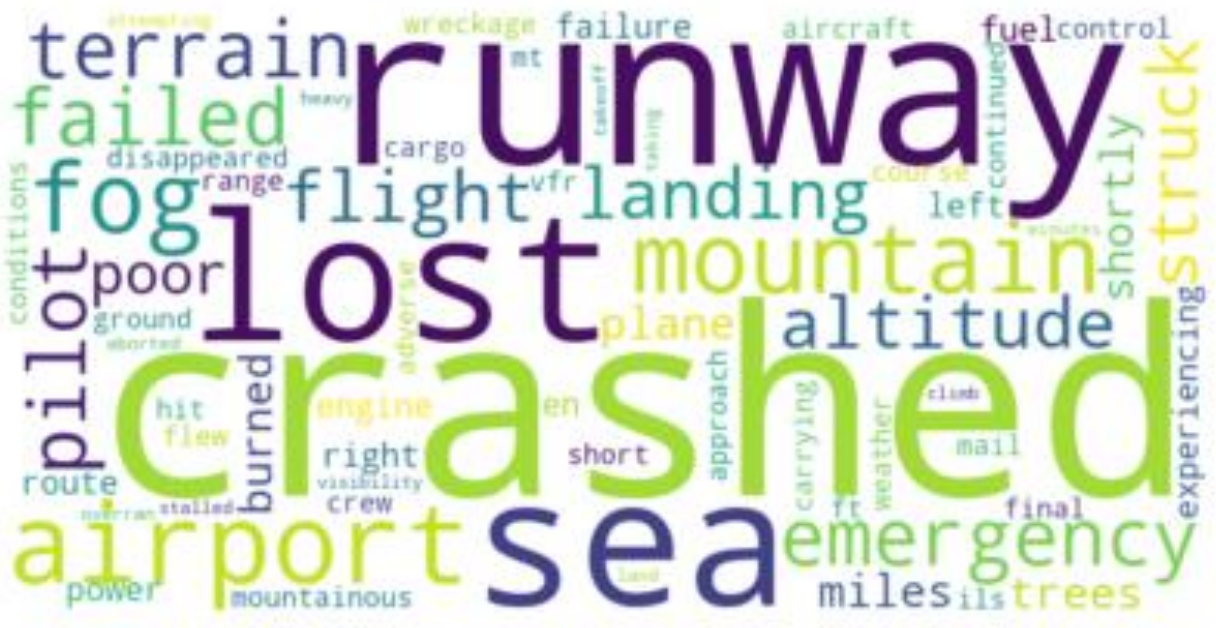

Fig. 6. Word Cloud of Topics for NMF Model

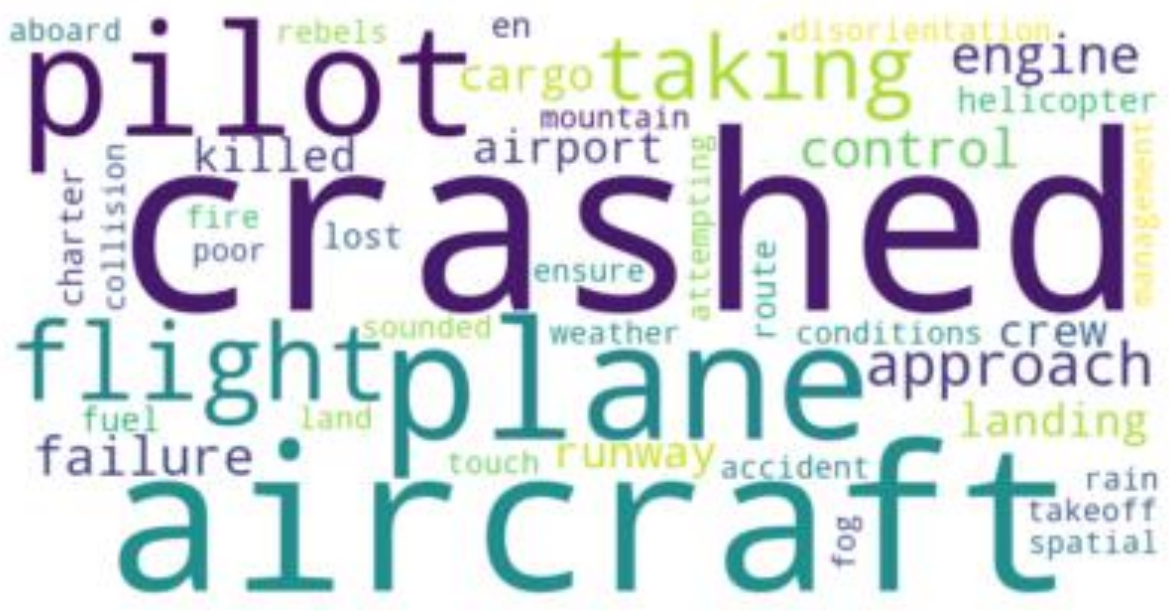

Fig. 7. Word Cloud of Topics for PLSA Model

### C. *Performance Metrics*

The model performance is evaluated using two key metrics that is coherence score and perplexity. The coherence score measures the interpretability of the topics, with higher values indicating more meaningful and human-understandable topics. Perplexity, on the other hand, assesses how well the model predicts the data, with lower values indicating better predictive performance.

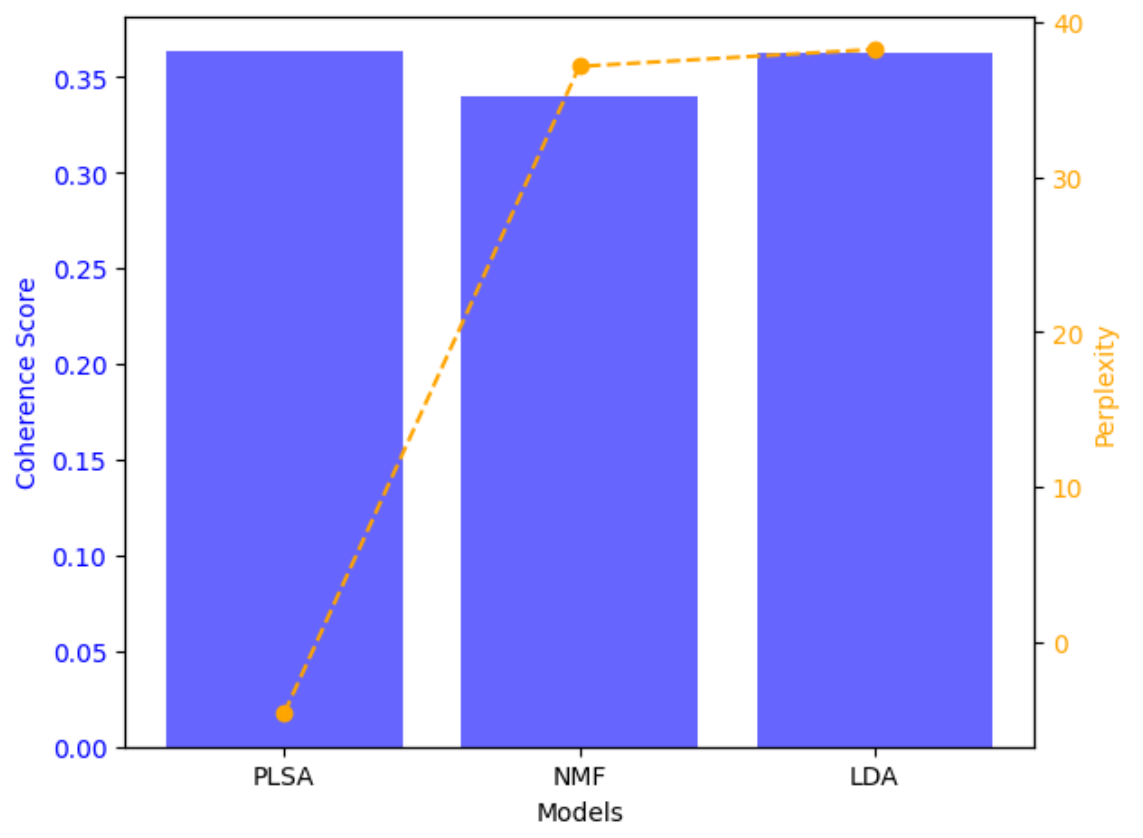

Fig. 8. Coherence Scores and Perplexity for Topic Model

PLSA achieved a coherence score of *0.32* and a perplexity value of *-4.6*, suggesting that while it provides better predictive performance, it has slightly less interpretability compared to NMF and LDA. NMF scored a coherence of *0.34* and a perplexity of *37.1*, producing more interpretable topics but with a higher perplexity value, indicating slightly reduced predictive accuracy. LDA, with the highest coherence score of *0.36* and the highest perplexity value of *38.2*, generated the most interpretable topics but demonstrated the poorest predictive performance among the three models. The coherence scores and perplexity values for each model are compared in Fig. 8. The blue bars represent the coherence scores, while the orange line shows the perplexity values for each model. Table II compares three topic modelling techniques LDA, NMF, and PLSA across various criteria, including topic interpretability, computational complexity, accuracy, strengths, weaknesses, and suitability for aviation safety data.

TABLE II. COMPARISON OF TOPIC MODELING TECHNIQUES (LDA, NMF, PLSA).

| **Criteria** | **LDA** | **NMF** | **PLSA** |
|---|---|---|---|
| **Topic Interpretability** | Topics are interpretable but overlap between themes may exist. | Topics are generally easy to interpret due to the non-negative constraints. | Some topics are harder to interpret due to their probabilistic nature. |
| **Computational Complexity** | Iterative Gibbs sampling can be computationally intensive. | The factorization process can be time-consuming for large datasets. | The Expectation-Maximization (EM) algorithm can be slower with larger datasets. |
| **Accuracy of Topic Representation** | Well-suited for uncovering hidden themes in the text. | Effective at representing distinct themes, especially with clearer, non-overlapping data. | Provides more probabilistic insights, but may not always offer clean, distinct topics. |
| **Strengths** | Ability to capture mixed-topic distributions in documents. | Good at extracting meaningful, non-overlapping topics. | Provides a detailed probabilistic framework, accounting for uncertainty. |
| **Weaknesses** | Can result in overlapping topics, which may reduce interpretability. | Sensitive to initial factors; may not perform well on highly complex datasets. | Can be challenging to interpret and computationally expensive due to the EM algorithm. |
| **Suitability for Aviation Safety Data** | Excellent for uncovering general themes, especially when documents have mixed topics. | Excellent for producing clear, interpretable topics from aviation safety narratives. | Good for probabilistic understanding, but may result in overlapping themes, making interpretation harder. |

## V. CONCLUSION

This study conducted an in-depth analysis of aviation safety records from 1908 to 2009, leveraging three topic modelling techniques LDA, NMF, and PLSA to extract themes from accident narratives spanning military, commercial, and private aviation sectors. By categorizing data based on operator type, the study uncovered unique patterns such as *pilot error*, *mechanical failures*, *weather conditions*, and *training-related issues* specific to military aviation.

The findings highlighted the strengths and limitations of each technique. LDA effectively identified overlapping thematic structures, making it suitable for datasets with mixed topics. NMF provided distinct and interpretable topics, aligning well with datasets requiring clear categorization. PLSA captured nuanced probabilistic relationships between terms but presented challenges in interpretability. Collectively, these techniques demonstrated the potential of topic modelling in deriving meaningful insights from unstructured aviation safety data.

The practical implications of this research extend to enhancing aviation safety protocols, guiding policy formulation, and optimizing operational strategies across different aviation sectors. By pinpointing key factors contributing to incidents, safety regulators and operators can prioritize targeted interventions and resource allocation.

Future work can expand upon these findings by incorporating additional variables such as environmental factors, aircraft types, and operational contexts to yield deeper insights. Exploring advanced methodologies, including neural topic models, could enhance the handling of large datasets and complex term-topic relationships. Finally, integrating topic modelling results with sentiment analysis and other text-mining techniques could enhance decision-making processes for aviation professionals. Such integration may lead to automated systems for incident reporting and risk assessment, ultimately strengthening aviation safety management practices.